\documentclass[a4paper,fleqn]{cas-dc}

\usepackage[authoryear,longnamesfirst]{natbib}
\usepackage{graphicx}
\usepackage{tabularx}
\usepackage{algorithmicx}
\usepackage{algpseudocode}
\usepackage[linesnumbered,ruled]{algorithm2e}

\def\tsc#1{\csdef{#1}{\textsc{\lowercase{#1}}\xspace}}
\tsc{WGM}
\tsc{QE}

\begin{document}
\let\WriteBookmarks\relax
\def\floatpagepagefraction{1}
\def\textpagefraction{.001}

\shorttitle{Temporal Shuffling for Defending Deep Action Recognition Models against Adversarial Attacks}    

\shortauthors{Jaehui Hwang, Huan Zhang, Jun-Ho Choi, Cho-Jui Hsieh, Jong-Seok Lee}  

\title [mode = title]{Temporal Shuffling for Defending Deep Action Recognition Models against Adversarial Attacks}  




\author[1]{Jaehui Hwang}[]




\author[2]{Huan Zhang}[]

\author[1]{Jun-Ho Choi}[]

\author[3]{Cho-Jui Hsieh}[]

\author[1]{Jong-Seok Lee\corref{cor1}}[
orcid=0000-0002-8038-1119
]
\renewcommand{\eadauthor}{J.-S. Lee}
\ead{jong-seok.lee@yonsei.ac.kr}
\cortext[cor1]{Corresponding author}

\affiliation[1]{organization={School of Integrated Technology, Yonsei University},
            country={Republic of Korea}}

\affiliation[2]{organization={Department of Computer Science, Carnegie Mellon University},
            country={USA},
}
            
\affiliation[3]{organization={Department of Computer Science, University of California},
            city={Los Angeles},
            country={USA}}




\begin{abstract}
Recently, video-based action recognition methods using convolutional neural networks (CNNs) achieve remarkable recognition performance.
However, there is still lack of understanding about the generalization mechanism of action recognition models.
In this paper, we suggest that action recognition models rely on the motion information less than expected, and thus they are robust to randomization of frame orders. Furthermore, we find that motion monotonicity remaining after randomization also contributes to such robustness.
Based on this observation, we develop a novel defense method using temporal shuffling of input videos against adversarial attacks for action recognition models. Another observation enabling our defense method is that adversarial perturbations on videos are sensitive to temporal destruction. To the best of our knowledge, this is the first attempt to design a defense method without additional training for 3D CNN-based video action recognition models.
\end{abstract}




\maketitle

\section{Introduction}\label{}
Human action recognition has been extensively researched with the improvement of deep neural networks. A key issue is how to effectively model the temporal motion patterns of actions with deep models. 3D convolutional neural networks (CNNs) performing three-dimensional (2D spatial and temporal) convolutional operations are the popular approaches for this, which achieved high recognition performance \citep{carreira2017quo,feichtenhofer2020x3d,feichtenhofer2019slowfast,ji20123d,tran2015learning,tran2019video,wang2018non, bertasius2021space, mazzia2022action}.

Deep learning models have shown remarkable performance in various areas.
To understand this success, researchers have explored which features the models learn and how the models generalize to data, especially for object recognition in images \citep{baker2018deep,geirhos2018imagenet,shetty2019not,wang2020high,xiao2020noise,zhu2017object}.
{Contrary to our expectation, it was found that trained object recognition models heavily rely on components that humans hardly exploit, such as high-frequency components \citep{wang2020high} and backgrounds \citep{xiao2020noise,zhu2017object}.}

\begin{figure}
	\centering
	\includegraphics[clip, trim=0cm 7cm 19cm 0cm, width=\columnwidth]{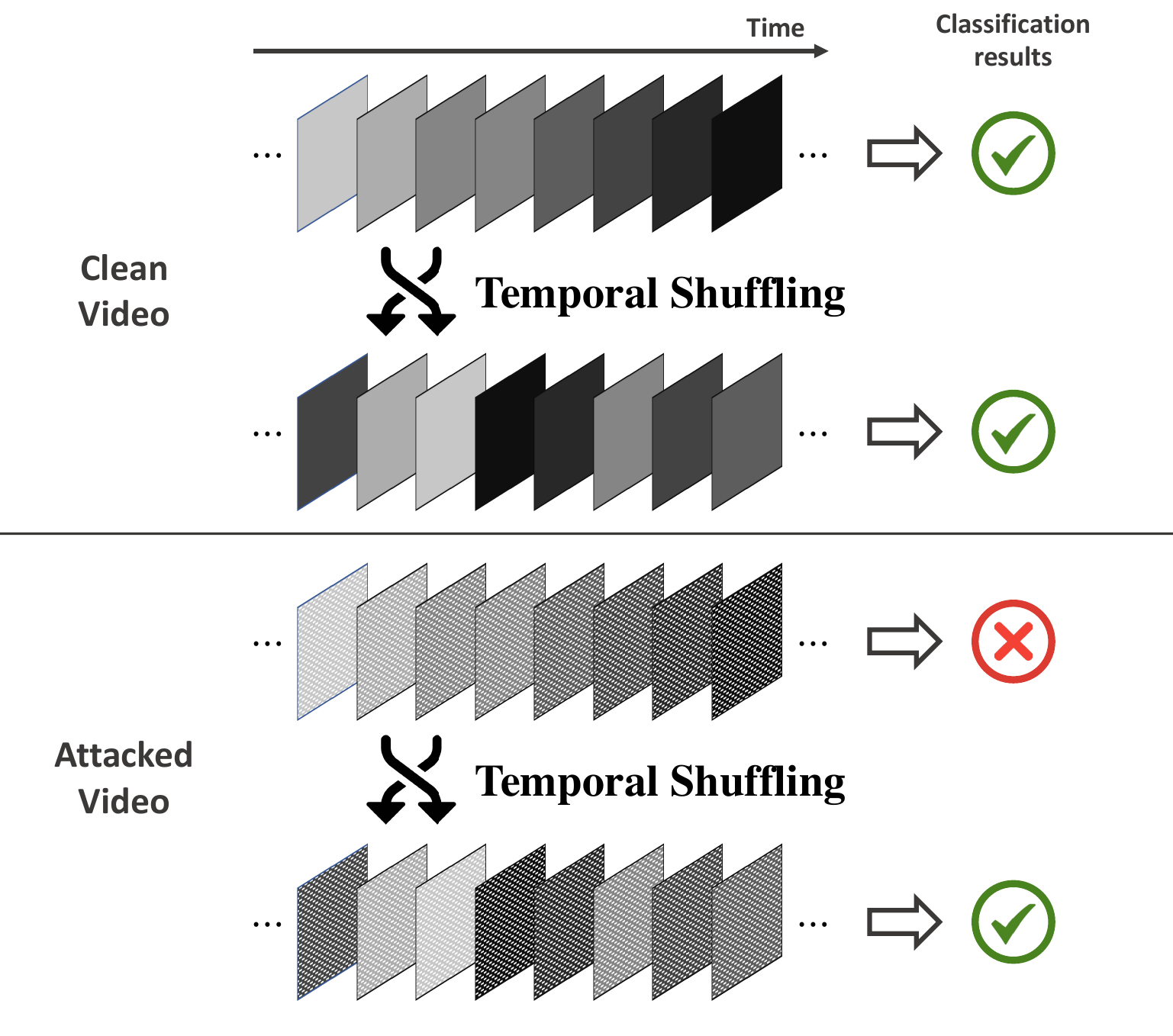}
	\caption{Temporal shuffling is not critical to the clean original video, but destroys adversarial perturbations attacking the action recognition model. We suggest a new defense method using this property.}
	\label{fig:teaser}
\end{figure}

Motivated by these findings and considering the aforementioned key issue of action recognition models, we pose the following question: \emph{Do trained action recognition models generalize by fully utilizing the motion pattern appearing in the input video?}
To answer this question, we analyze the dependence of action recognition models on spatial and temporal information through various experiments.
We find that the models depend on the motion information only partly and are robust to motion variation.
Even when the order of video frames is randomized and the temporal information is modified, the models maintain relatively high performance.

We exploit this property to propose a novel method to defend against adversarial attacks on action recognition models. 
In particular, we show that unlike videos, adversarial perturbations on action recognition models are easily neutralized by randomization of the frame order, which facilitates our defense method together with the aforementioned property (Figure~\ref{fig:teaser}).
The main contributions of this work can be summarized as follows.

\begin{itemize}
\item
	We discover that different from our expectation, state-of-the-art action recognition models rely on motion information only marginally, because they significantly depend on spatial information such as objects and backgrounds. In addition, we show that the models generalize well for varying motion speed if the monotonicity of motion is preserved. These characteristics make the models robust to temporal destruction such as randomization of the frame order.
	\item
	We explore the importance of temporal orders of adversarial perturbations that attack action recognition models, and find that the perturbations are sensitive to temporal destruction. Two factors behind this observation are identified, namely, the importance of the temporal pattern appearing in an adversarial perturbation and the importance of the proper temporal location of a frame perturbation.
	\item
	Based on these findings, we suggest a novel defense method, called temporal shuffling, on action recognition models. To the best of our knowledge, it is the first attempt of adversarial defense without additional training for 3D CNN-based action recognition models. 
\end{itemize}

\section{Related work}

\subsection{Action recognition}

Early studies on action recognition often used recurrent neural networks to model sequential features \citep{donahue2015long}. However, the current state-of-the-art approach is to use 3D CNNs that can extract features considering both spatial and temporal dimensions. I3D \citep{wang2018non} inflates 2D convolutional kernels in ResNet to 3D. In the interaction-reduced channel separated network (ir-CSN) \citep{tran2019video}, a kernel factorization technique is used to reduce computational complexity. SlowFast \citep{feichtenhofer2019slowfast} is composed of two streams of 3D CNNs receiving two types of video data with different temporal resolutions. X3D \citep{feichtenhofer2020x3d} reduces complexity by finding optimal design factors such as temporal activation size, spatial resolution, etc. Nowadays, transformer-based action recognition models \citep{bertasius2021space, mazzia2022action} also exist, but attacks on such models have been rarely attempted.

\subsection{Model generalization}

There exist attempts to explain the remarkable generalization capability of deep neural networks for object recognition. CNNs generalize various correlations between images and their class labels \citep{baker2018deep,geirhos2018imagenet,shetty2019not,wang2020high,xiao2020noise,zhu2017object}. However, some of the correlations are different from human expectations. For instance, people hardly find meaningful information from high frequency components in images, but CNN models tend to rely on them \citep{wang2020high}. In addition, many models exploit backgrounds of images for classification \citep{xiao2020noise,zhu2017object}. However, analysis on the generalization capability of video-based action recognition models is rarely found in literature. Since video data have the temporal dimension unlike images, the findings in object recognition are not directly applied to action recognition. 
We present our results on this topic, particularly focusing on the temporal dimension.

\subsection{Adversarial attack and defense}

Many deep models have shown severe vulnerability to adversarial attacks, which fool classification models by perturbing input data. 
A popular approach is gradient-based optimization of perturbations, such as the fast gradient sign method (FGSM) \citep{goodfellow2014explaining} and its iterative version, I-FGSM \citep{kurakin2016adversarial}.
There exist a few attack methods on action recognition models. The sparse attack \citep{wei2019sparse} aims to perturb only a few frames to attack LSTM-based models. The flickering attack \citep{pony2021over} changes the overall color of each frame. The one frame attack \citep{hwang2021just} inserts a perturbation to only one frame by exploiting the structural vulnerability of the given model. \citet{li2018adversarial} investigated the problem of applying attacks on real-time video classification systems.

Several defense methods have been developed for object recognition. Heuristic approaches include random resizing \citep{xie2018mitigating}, JPEG compression \citep{dziugaite2016study}, etc. Recently, defense methods that theoretically certify robustness have been studied, such as randomized smoothing \citep{cohen2019certified} and denoised smoothing \citep{salman2020denoised}.

A few studies suggested adversarial defense methods for action recognition models. \citet{anand2020adversarial} applied local gradient smoothing to an adversarial patch attack on optical flow-based action recognition models. MultiBN \citep{lo2021defending} is a batch normalization method for improving the robustness of the models. However, MultiBN needs additional training, while video-based models are hard to train due to the high computational complexity. In addition, these studies did not consider recent state-of-the-art models.


\section{Influence of temporal changes of videos}
\label{sec: 3}

In this section, we investigate how important temporal motion cues contained in videos are for action recognition. 
We first show that action recognition models are robust to destruction of temporal information.
Then, experiments to explain such robustness are conducted.

\subsection{Robustness to temporal destruction}
\label{sec: 3-1}

To properly evaluate the robustness of action recognition models against the destruction of temporal information, we examine the recognition performance when the temporal information of input videos is destroyed.
Two types of destruction are considered: uniformization of video frames and randomization of frame orders.
The former completely removes temporal cues in a video, while the latter disturbs natural motion.
\\[-0.5\baselineskip]

\noindent \textbf{Uniformizing video frames.} 
Let $X=\{x(1), ... , x(T)\}$ denote an original video having $T$ frames. Then, its uniformized version is defined as $X' = \{x(i), x(i), ..., x(i)\}$, where $i \in [1, ... , T]$.
In other words, the uniformized video is visually static without any motion.
In the experiments, we try all possible values of $i$ (i.e., $i$=1 to $T$).
\\[-0.5\baselineskip]

\noindent \textbf{Randomizing frame orders.} 
In this case, frames in a video are randomly permuted. To control the amount of destruction, random permutation occurs only within each chunk composed of $N$ frames. To be more specific, the given video is divided into multiple disjoint chunks with $N$ successive frames, and the frame order in each chunk is randomized independently. For instance, with $N$=4, the first chunk can be changed from $\{x(1),x(2),x(3),x(4)\}$ to $\{x(3),x(1),x(4),x(2)\}$.
For comparison, spatial destruction is also conducted in a similar way, i.e., the order of the rows in each group of $N$ rows is randomly permuted. Note that the random order is kept the same for all frames.
\\[-0.5\baselineskip]

\noindent \textbf{Experimental details.} Kinetics-400 \citep{kay2017kinetics} is a popular large-scale dataset in action recognition.
We randomly choose ten videos in each class from the test set of Kinetics-400. Among them, we use the 1900 videos that are correctly classified by all models for fair comparison across different models. We employ four state-of-the-art action recognition models, including 
I3D \citep{wang2018non}, SlowFast \citep{feichtenhofer2019slowfast}, ir-CSN \citep{tran2019video}, and X3D \citep{feichtenhofer2020x3d}. 
We use the pre-trained models on Kinetics-400 from MMAction2 \citep{2020mmaction2}.
Among the multiple versions of SlowFast and X3D, we choose 8$\times$8 SlowFast and X3D-M, respectively.
These video dataset and models are used throughout this paper.
Note that X3D receives videos having 16 frames, while 32 frames are used for the others.
Thus, the chunk size for randomization is set to $N \in \{4, 8, 16\}$ for X3D and $N \in \{4, 8, 16, 32\}$ for the other models.
\\[0.5\baselineskip]

\newcolumntype{b}{>{\hsize=1\hsize\centering\arraybackslash}X}
\newcolumntype{s}{>{\hsize=0.85\hsize \centering\arraybackslash}X}
\newcolumntype{a}{>{\hsize=0.9\hsize \centering\arraybackslash}X}
\newcolumntype{l}{>{\hsize=1.2\hsize \centering\arraybackslash}X}
\newcolumntype{k}{>{\hsize=1.3\hsize \centering\arraybackslash}X}

\begin{table}[t]
	\centering
         \caption{Classification accuracy of uniformized videos.}
	\begin{tabularx}{\columnwidth}{b|bbbb}
		\toprule
		Model&I3D
            &SlowFast
            &ir-CSN
            &X3D
            \\
		\midrule
		Accuracy&69.1\%&69.4\%&62.3\%&67.4\% \\
		\bottomrule
	\end{tabularx}
	\label{table:clean_uniform}
	\centering
\end{table}

\begin{figure}[t]
	\centering
	\includegraphics[width=0.85\columnwidth]{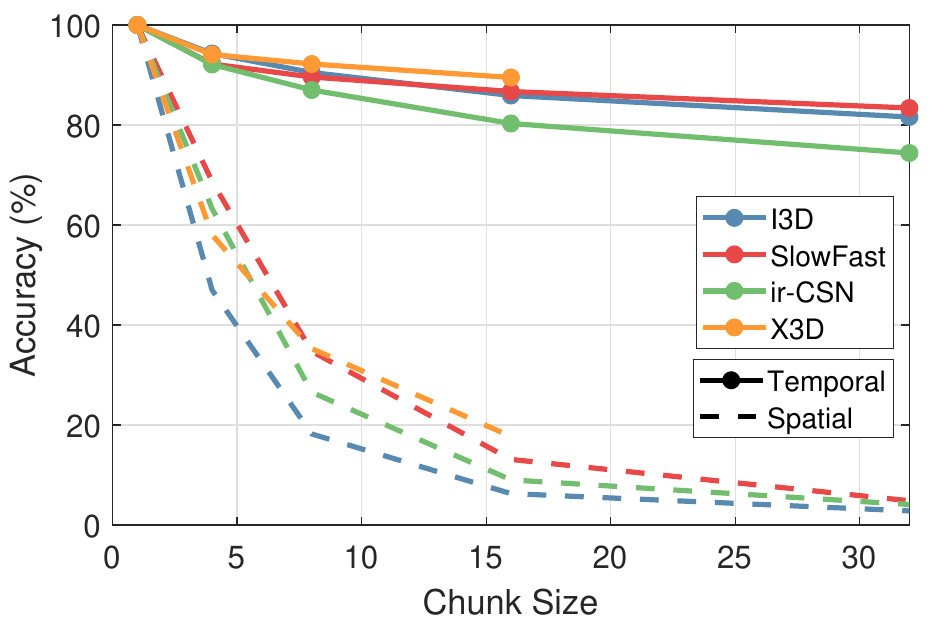}
	\caption{Classification accuracy of the videos with randomizing the order of frames with respect to the chunk size (`Temporal'). Results with randomizing the order of rows  are also shown for comparison (`Spatial').}
	\label{fig:clean_randomizing}
\end{figure}

\noindent \textbf{Results.} Table \ref{table:clean_uniform} shows the accuracy for the uniformized videos.
Note that the accuracy for the original videos is 100\%. It is quite surprising that over 60\% of the uniformized videos are classified correctly by all models even though there are no motion cues in the videos. This suggests that the models can extract useful features from spatial cues.

Figure \ref{fig:clean_randomizing} shows the results for randomization of frame orders.
Note that the case with $N$=1 is for the original videos, for which the accuracy is 100\%.
As the chunk size becomes larger, the motion information is more distorted and thus the accuracy decreases. However, the accuracy drops are rather mild; even when the randomization is performed over the whole range (i.e., $N$=16 for X3D and 32 for the other models), the accuracy is higher than that for uniformized videos.
Furthermore, the temporal changes degrade the recognition performance much less than the spatial changes. These results indicate that the action recognition models are fairly robust to temporal changes. In the following, we explore the reasons of this robustness further.

\subsection{Dependence on spatial information}
\label{sec: 3-2}

First, we explain the robustness of the models against temporal destruction in the viewpoint of dependence of the models on spatial and temporal information.
\\[-0.5\baselineskip]

\begin{figure*}[t]
	\centering
	\includegraphics[width=0.7\textwidth]{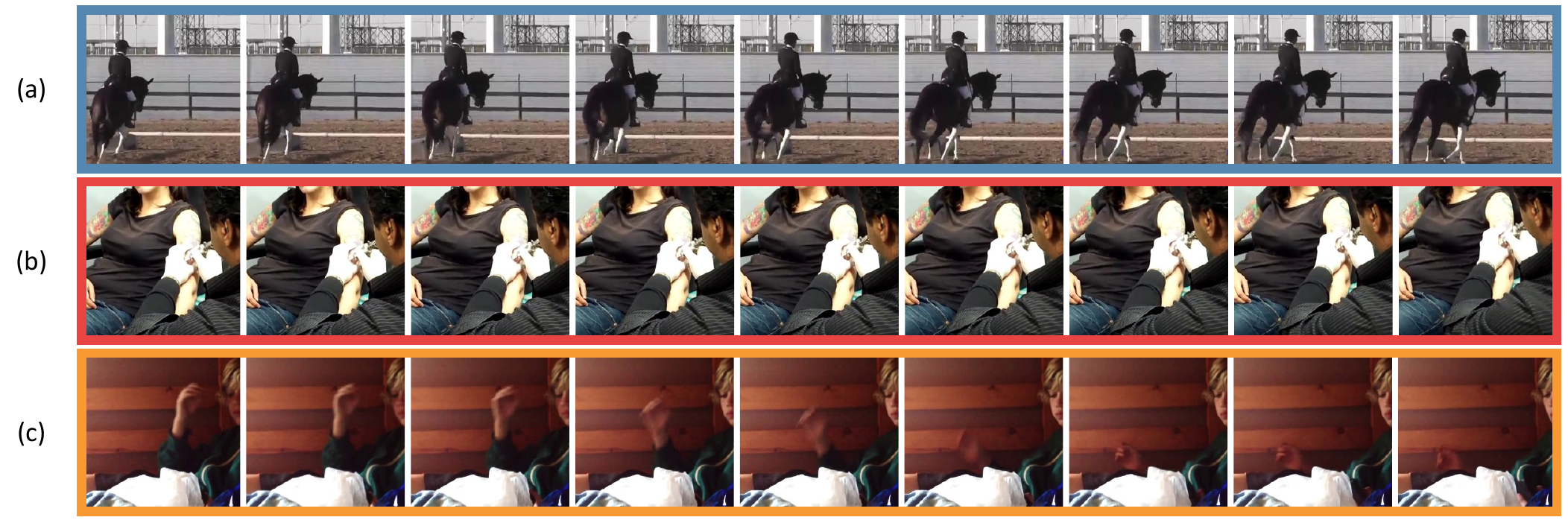}
    \includegraphics[width=0.28\textwidth]{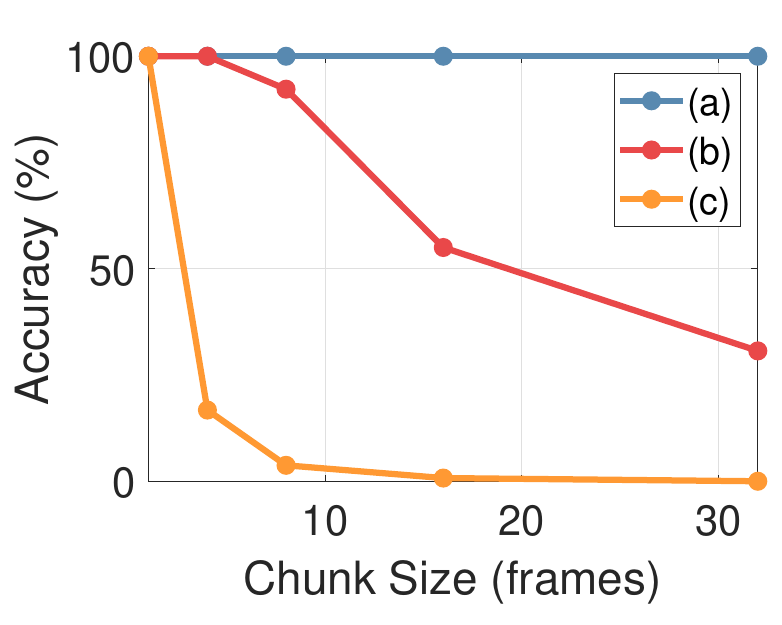}
	\caption{(Left) Example videos having different amounts of class-relevant spatial cues. (Right) Classification accuracy of the three videos under randomization of frame orders.
	}
	\label{fig:data_type}
\vspace{-1em}
\end{figure*}

\noindent \textbf{Representative cases.} We consider three categories of videos: (1) videos containing informative backgrounds, moving objects (whole bodies, hands, etc.), and their motion, (2) videos containing informative cues in objects and motion (but not in backgrounds), and (3) videos containing informative cues only in motion. Here, being informative means providing useful information for classification.
The left panel of Figure \ref{fig:data_type} shows a representative video in each category.
In Figure \ref{fig:data_type}(a), there is clear information on the background (horse-breeding farm) and objects (horse and rider), which are relevant to the true class label, ``Riding or walking with a horse.''
Figure \ref{fig:data_type}(b) with a true class label ``Getting a tattoo'' contains class-relevant information on the object (tattoo machine). In the third case in Figure \ref{fig:data_type}(c), both background (wooden wall) and object (hand) are not so specific to the class label ``Eating chips.''
The right panel of Figure \ref{fig:data_type} shows the accuracy of each of the three videos for 100 random permutations of frame orders, which is averaged over the four models. Since the first video has rich spatial information, it is always correctly classified even under the randomization. The second video shows moderately robust performance thanks to the class information in the object. On the other hand, the third video, which has little class-related spatial information, is vulnerable to temporal destruction.
These results explain the unexpectedly high accuracy shown in Table \ref{table:clean_uniform} and Figure \ref{fig:clean_randomizing}. The models heavily rely on spatial cues whenever available, thus the effect of temporal destruction is not significant.
\\[-0.5\baselineskip]

\begin{figure}[t]
    \centering
    \includegraphics[clip, trim= 0cm -4cm 0 0, width=0.22\columnwidth]{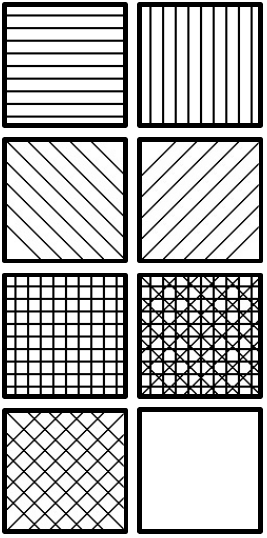}\hspace{4mm}
    \includegraphics[clip, trim= 0cm -4cm 0 0,width=0.22\columnwidth]{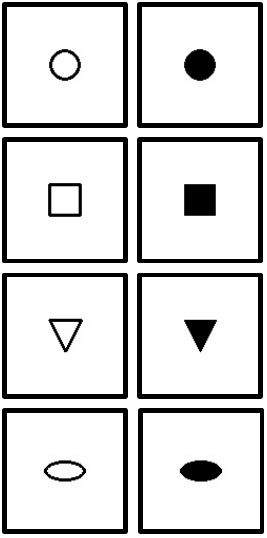}\hspace{4mm}
    \includegraphics[clip, trim= 0cm -3cm 0 0,width=0.22\columnwidth]{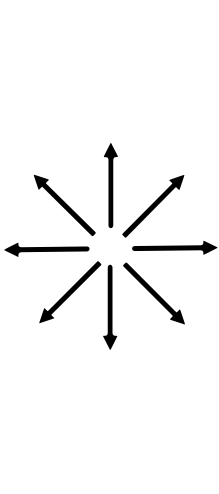}
    \vspace{-4em}
    \caption{Patterns of backgrounds (left), objects (middle), and motions (right) used to generate datasets for the toy experiment.}
    \label{fig:toy_feature}
\end{figure}

\begin{figure}[t]
    \centering
    \includegraphics[width=\columnwidth]{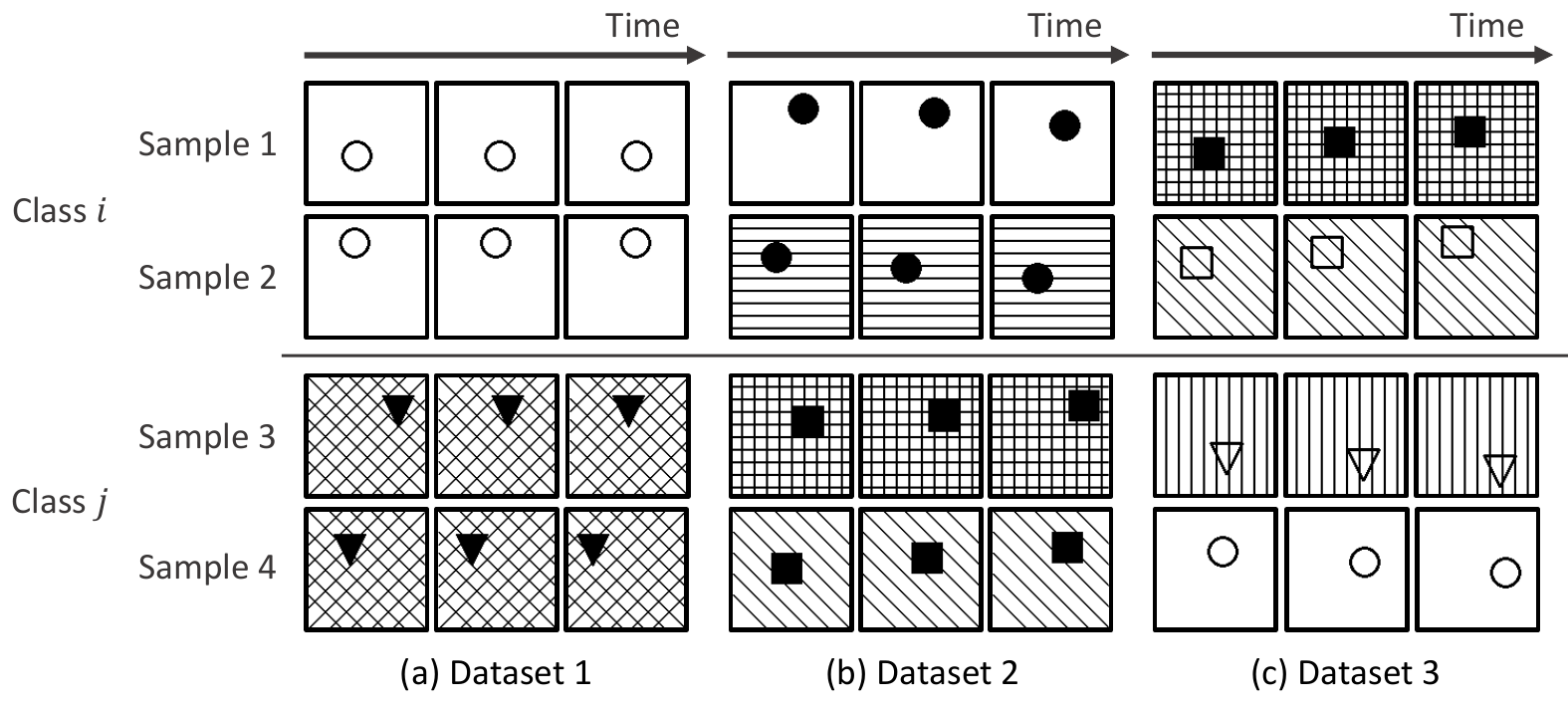}
    \caption{Samples from the toy datasets. Each row represents a distinct sample. Note that the samples in the upper (or lower) two rows are from the same class in each dataset.}
    \label{fig:dataset-ex}
\end{figure}

\begin{figure}[t]
    \centering
    \includegraphics[width=0.85\columnwidth]{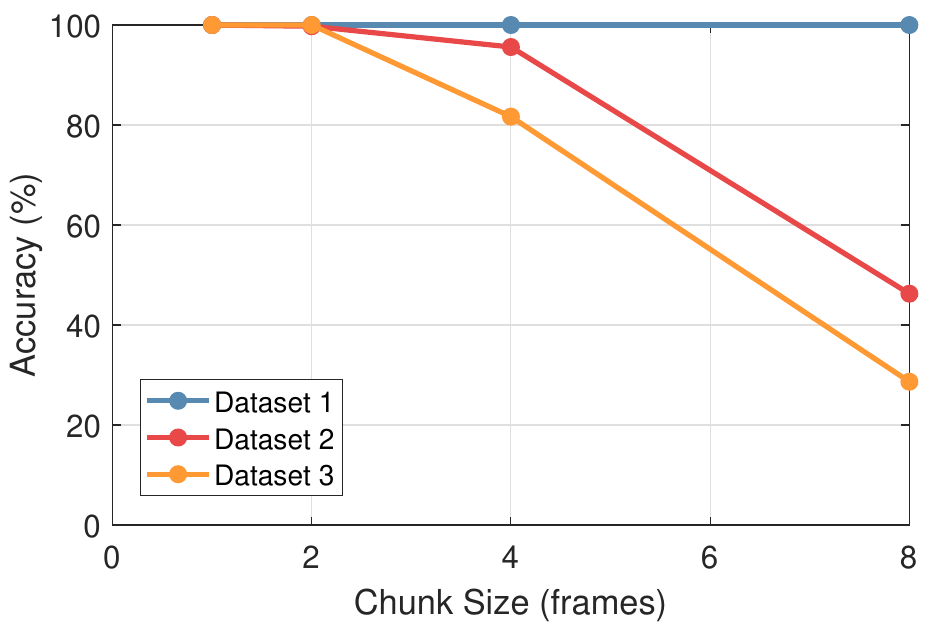}
    \caption{Classification accuracy for the toy experiment under randomization of frame orders.}
    \label{fig:toy_randomizing}
\end{figure}

\noindent \textbf{Toy experiment.} 
We verify the above explanation further in a controlled manner through a toy experiment.
We generate eight-frame-long videos containing a simple object moving in a certain direction on a simple background.
The dataset is available at \url{https://github.com/J-H-Hwang/temporal_shuffling}.
We consider eight types of background, eight types of object, and eight motion directions (Figure \ref{fig:toy_feature}) in order to pose an eight-class classification problem in each of three datasets.
In Dataset 1, a combination of one type of background, one type of object, and one motion direction is assigned to each class, so that each of the background, object, and motion in a video contains class-relevant information. 
Figure \ref{fig:dataset-ex} (a) shows two samples for \{background: empty, object: empty circle, motion: right\} (upper rows) and two samples for \{background: diagonal crosshatch, object: full triangle, motion: left\} (lower rows).
In Dataset 2, a combination of one type of object and one motion direction is assigned to each class, while the background is randomly chosen. 
Figure \ref{fig:dataset-ex} (b) shows two samples for \{object: full circle, motion: down and right\} (upper rows) and two samples for \{object: full square, motion: up and right\} (lower rows), where the background is chosen randomly in each sample.
In Dataset 3, each class corresponds to one of the eight motion directions, and the background and object are chosen randomly. 
In Figure \ref{fig:dataset-ex} (c), we show two samples for \{motion: up and right\} (upper rows) and two samples for \{motion: down and right\} (lower rows) with random objects and backgrounds. 
The motion speed is constant in each video between 3 and 5 pixels per frame. Each dataset contains 1200 training videos and 400 test videos. For each dataset, we train a 3D CNN model based on ResNet18 using the Adam algorithm \citep{adam}, where the 3$\times$3 convolutional layers are inflated to 3$\times$3$\times$3 convolutional layers.
Figure \ref{fig:toy_randomizing} shows the accuracy under randomization of frame orders for the three datasets.
When the spatial cues in videos are related to the class labels (i.e., Dataset 1), the models show robust performance.
The largest accuracy drop is observed in Dataset 3 where only temporal cues are class-relevant.
These results also support that the models are trained to exploit spatial cues effectively, so they become robust to temporal destruction.

\subsection{Robustness to motion variation}
\label{sec: 3-3}

We note that in Figure \ref{fig:toy_randomizing}, the accuracy at $N$=8 (i.e., randomization over all frames) for Dataset 3 (with no spatial cues) is 28.7\%, which is still significantly higher than random chance (1/8=12.5\%).
This suggests that even under complete randomization, certain motion cues useful for recognition still remain.

\begin{figure}[t]
	\centering
	\includegraphics[width=\columnwidth]{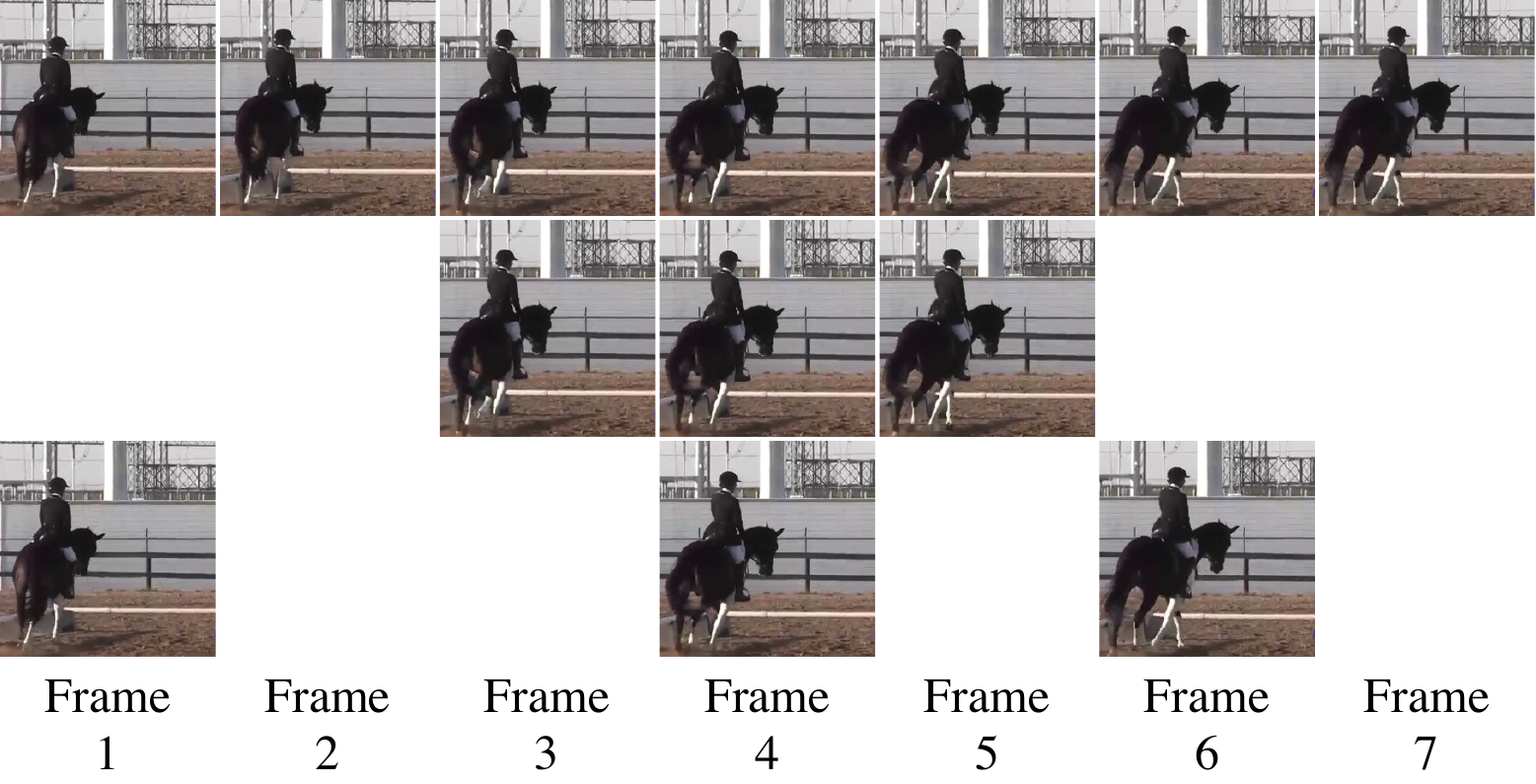}
	\caption{(Top) Original video frames. (Middle) Frames are selected consecutively. Motion information within these frames remains the same to that in the original video. (Bottom) Frames are selected irregularly. The original motion data is partly corrupted but the monotonicity is preserved.}
	\label{fig:concept}
 \vspace{-1em}
\end{figure}

We hypothesize that \textit{monotonicity} of motion can be informative for recognition even under temporal randomization. The second row in Figure \ref{fig:concept} shows a part of the original video (the first row), which locally preserves the motion information. The third row, where frames are chosen at an irregular interval, contains corrupted motion but the monotonicity (i.e., the horse walking from left to right) is maintained. Randomization of frame orders may produce this type of irregular but monotonic frame orders, from which a recognition model can still be able to extract useful motion features.

For the randomized frame orders obtained by `Randomizing frame orders' in Section \ref{sec: 3-1}, we calculate the ratio of monotonic frame orders in the viewpoint of the first convolutional layer of a model. For example, consider a randomized order [2, 3, 4, 1, 5]. When the first convolutional layer has a temporal kernel size of 3 with a stride of 1 and zero-padding is used, the kernel processes the following five sets of frames: [2, 3], [2, 3, 4], [3, 4, 1], [4, 1, 5], and [1, 5]. Since the first, second, and the last sets contain monotonic orders, the ratio of monotonic frame orders is 3/5=60\%.

\begin{figure}[t]
	\centering
        \includegraphics[width=0.85\columnwidth]{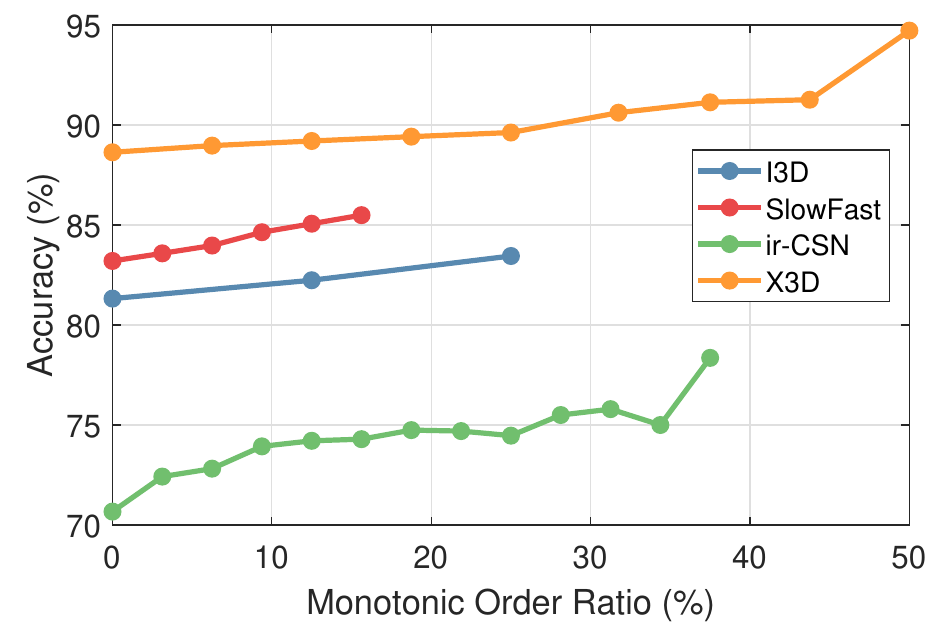}
        \caption{Classification accuracy under randomization of frame order with respect to the ratio of monotonic orders.}
        \label{fig:tend_}
\end{figure}

Figure \ref{fig:tend_} shows the recognition accuracy for the randomized videos with respect to the ratio of monotonic frame orders when $N$=32.
Note that the possible values of the ratio are different among the models because of different temporal strides in the first convolutional layers.
In SlowFast and ir-CSN, the monotonic order ratio can be 1/32, ..., 32/32 because the temporal size of the input is 32 and the stride is 1.
The ratio of X3D can be 1/16, ..., 16/16 because the stride is 1 but the temporal size of the input is 16.
In the case of I3D, the temporal size of the input is 32 and the stride of the first convolutional layer is 2, which would result in possible ratios of 1/16, ..., 16/16; however, since the subsequent pooling layer having a stride of 2 drops every other frame, effective monotonic order ratios of I3D become 1/8, ..., 8/8.
In Figure \ref{fig:tend_}, the x-axis values of the data points correspond to these values.
Note that it is probabilistically unlikely to obtain a randomized video having a high monotonic order ratio.
Thus, no or too few videos correspond to high ratio values, for which reliable accuracy cannot be obtained.
We exclude such cases, which is why the data points exist only up to certain ratio values in the figure (e.g., only 0/8, 1/8, and 2/8 for I3D).
From the figure, it is observed that the accuracy increases as the ratio increases, which supports our hypothesis of usefulness of motion monotonicity.

\section{Influence of temporal changes of \\ adversarial perturbations}
\label{sec: 4}

In this section, we examine the effect of temporal changes on adversarial perturbations that are added to original videos to cause misclassification. We show that unlike the clean videos, adversarial perturbations are sensitive to temporal changes, and investigate reasons of such sensitivity.

\subsection{Impact of temporal destruction}
\label{sec: 4-1}

\begin{figure}[t]
 	\centering
        \includegraphics[width=0.85\columnwidth]{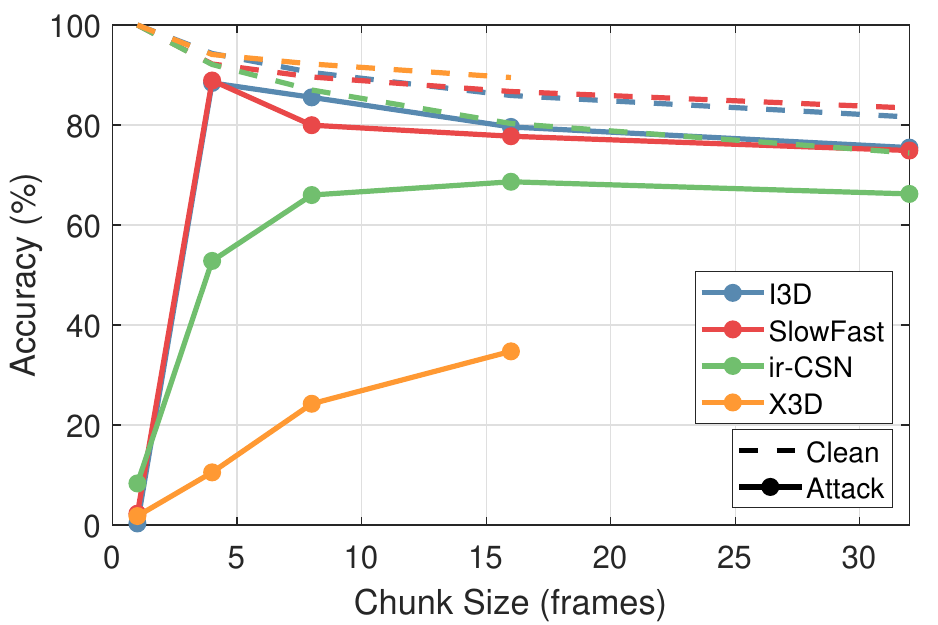}
        \caption{Classification accuracy of the attacked videos by I-FGSM ($\epsilon$=4/255) after randomization of the orders of frames.}
        \label{fig:attack_random}
\end{figure}

We use I-FGSM \citep{kurakin2016adversarial}, which is one of the strong image attack methods, to obtain a 3D perturbation having the same size to the given video to attack videos.
Let $P^0=\{p^0(1), ... , p^0(T)\}$ denote an initial perturbation whose elements are set to zero.
The attack aims to find a perturbation $P^M$ iteratively (with $M$ iterations), where the perturbed video by adding $P^M$ to the original video is misclassified. I.e., $f(X+P^M) \neq y$ while $f(X) = y$, where $f(\cdot)$ is the target action recognition model.
The I-FGSM update rule finds $P^{m+1}$ from $P^m$ iteratively:
\begin{multline}
    \widetilde{P}^{m+1} = 
    \mathrm{Clip}_{-X,255-X} \Big( \\
    P^m+\frac{\epsilon}{M}~\mathrm{sgn}\big( \nabla_{P^m} J(X + P^m, y \big) \Big),
\label{eq:ifgsm}
\end{multline}
\begin{equation}
    P^{m+1} = \mathrm{Clip}_{-\epsilon, \epsilon} (\widetilde{P}^{m+1}),
\end{equation}
where $\epsilon$ is the $L_\infty$-bound of the perturbation, $\mathrm{sgn}(\cdot)$ is the sign function, $\nabla J(A,y)$ is the gradient of the loss $J(A,y)$, and
$\mathrm{Clip}_{c, d}(B) = \mathrm{min} \big( \mathrm{max} (B, c ), d \big)$,
which operates as an element-wise function if $c$ and $d$ are matrices.
After attacking the videos by I-FGSM with $M$=30, we randomize the orders of frames of the videos.

Figure \ref{fig:attack_random} shows the accuracy of the attacked videos under randomization of frame orders.
Without randomization ($N$=1), the attack successfully fools the models, yielding accuracies of 0.3\%, 2.3\%, 8.3\%, and 1.8\% for I3D, SlowFast, ir-CSN, and X3D, respectively. However, the temporal changes by randomization significantly reduce the effect of the attack. Except for X3D, the accuracies are over 50\% for all values of $N$. In particular, the attack is almost completely neutralized for I3D and SlowFast, achieving the accuracies close to those for the randomized clean videos. 

Below, we explore why temporal destruction can destroy adversarial attacks. 
We examine the importance of the temporal patterns in adversarial perturbations.
Then, we show that the frame locations of perturbations are important.

\subsection{Importance of temporal patterns}
\label{sec: 4-2}

To study the importance of the temporal patterns of adversarial perturbations, we test the efficacy of adversarial perturbations ignoring temporal patterns in two ways.
First, we set a constraint that adversarial perturbations are static over frames to exclude any temporal pattern completely.
Second, we generate adversarial perturbations without considering the temporal dimension, which still vary over frames but are not optimized in the temporal dimension.
\\[-0.5\baselineskip]

\noindent \textbf{Case 1: Generating static perturbations.} 
We introduce a static perturbation $P=\{p,p,...,p\}$, i.e., the same perturbation $p$ is added to all video frames. To update the perturbation $p$, we employ a modified I-FGSM update rule as follows:
\begin{multline}
    \widetilde{p}^{m+1} = 
    \mathrm{Clip}_{-min_t(X),255-max_t(X)}\Big(\\
    p^m+\frac{\epsilon}{M}~\mathrm{sgn}\big( \nabla_{p^m} J(X + \{p^m,p^m,...,p^m\}, y \big) \Big),
\label{eq:ifgsm}
\end{multline}
\begin{equation}
    p^{m+1} = \mathrm{Clip}_{-\epsilon, \epsilon} (\widetilde{p}^{m+1}),
\end{equation}
where $min_t(X)$ and $max_t(X)$ are the minimum and maximum values of $X$ through the time axis, respectively.
Thus, any temporal pattern does not exist in the perturbation.
\\[-0.5\baselineskip]

\newcolumntype{b}{>{\hsize=1\hsize\centering\arraybackslash}X}
\newcolumntype{s}{>{\hsize=1\hsize \centering\arraybackslash}X}
\renewcommand{\arraystretch}{1}
\begin{table}[t]
	\centering
         \caption{Classification accuracy of the videos that are perturbed by different attack methods. `Default' means the regular I-FGSM ($\epsilon$=4/255).}

	\begin{tabularx}{\columnwidth}{b|ssss}
		\toprule
		&I3D 
            &SlowFast
            &ir-CSN
            &X3D
            \\
		\midrule
		Default&0.3\%&2.3\%&8.3\%&1.8\%\\
		Case 1&17.9\%&27.5\%&26.6\%&8.3\%\\
		Case 2&1.0\%&5.7\%&11.8\%&2.4\%\\
		\bottomrule
	\end{tabularx}
	\label{table:temporal_pattern}
	\centering
\end{table}

\noindent \textbf{Case 2: Generating perturbations without considering the temporal dimension} 
In this case, we generate a multi-frame (3D) perturbation from one-frame (2D) perturbations that are computed individually for each frame. For this, we adopt the One Frame Attack (OFA) method proposed by \cite{hwang2021just}. In other words, we update only the one-frame perturbation ($p(i)$) within a sequence of multi-frame perturbation ($p(1), ..., p(T)$) using a modified I-FGSM rule, as expressed by the following equations:
\begin{multline}
    \widetilde{p}(i)^{m+1} = 
    \mathrm{Clip}_{-x(i),255-x(i)}\Big(p^m(i)+\frac{\epsilon}{M}~\mathrm{sgn}\big(\\
     \nabla_{p^m(i)} J(X + \{p^0(1),...,p^m(i),...,p^0(T)\}, y \big) \Big),
\label{eq:ifgsm}
\end{multline}
\begin{equation}
    p(i)^{m+1} = \mathrm{Clip}_{-\epsilon, \epsilon} (\widetilde{p}(i)^{m+1}),
\end{equation}
where $x(i)$ represents the $i$-th frame of the input $X$. Note that this approach generates one-frame perturbations independently without considering perturbations in other frames. Consequently, although the resulting 3D perturbation exhibits temporal changes, it does not exploit the temporal dimension to attack the model.
\\[-0.5\baselineskip]


\noindent \textbf{Results.} Table \ref{table:temporal_pattern} shows the accuracy for the two cases.
They always yield higher accuracy (i.e., lower attack success rates) than the unrestricted I-FGSM optimized in both spatial and temporal dimensions.
The number of variables that are optimized in the perturbation is $T$ times larger in Case 2 than in Case 1 because frame-wise perturbations are generated and merged in Case 2 while a temporally static perturbation is used in Case 1.
In other words, Case 2 has higher degrees of freedom in optimizing the perturbations than Case 1, and thus Case 2 shows lower accuracy than Case 1.
These results demonstrate that temporal patterns of adversarial perturbations are important for successful attacks. Nevertheless, there exist some gaps between significant removal of the effect of attacks in Figure \ref{fig:attack_random} and partial reduction of the effect of attacks in Table \ref{table:temporal_pattern}, suggesting that importance of temporal patterns is not a complete explanation for the sensitivity of adversarial perturbations to temporal destruction.

\newcolumntype{b}{>{\hsize=1.6\hsize\centering\arraybackslash}X}
\newcolumntype{s}{>{\hsize=0.85\hsize \centering\arraybackslash}X}
\renewcommand{\arraystretch}{1}

\begin{table}[t]
	\centering
         \caption{Classification accuracy of the attacked videos by I-FGSM ($\epsilon$=4/255) and their  uniformized versions.}

	\begin{tabularx}{\columnwidth}{b|ssss}
		\toprule
		&{I3D}
            &{\shortstack[lb]{SlowFast}}
            &{\shortstack[lb]{ir-CSN}}
            &{X3D}
            \\
		\midrule
        {Attacked}&{0.3\%}&{2.3\%}&{8.3\%}&{1.8\%} \\
		{{\shortstack[lb]{+Uniformized}}}&{61.1\%}&{60.4\%}&{52.9\%}&{45.9\%}
            \\
		\bottomrule
	\end{tabularx}
        \label{table:attack_uniform}
	\centering
\end{table}

\begin{figure}[t]
    \centering

	\includegraphics[width=0.31\columnwidth]{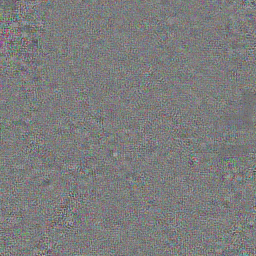}
	\includegraphics[width=0.31\columnwidth]{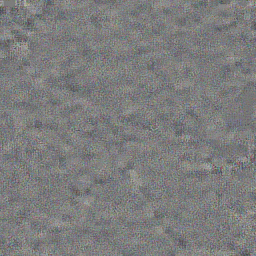}
	\includegraphics[width=0.31\columnwidth]{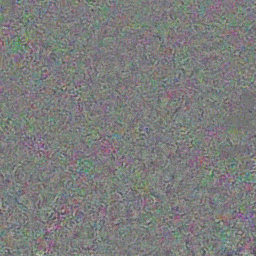}
	\caption{Adversarial perturbations on different frame locations: first, 12th, and 31st frame locations. Magnification by $\times$20 is applied for visualization.}
	\label{fig:sample_pert}
\end{figure}

\subsection{Importance of temporal locations}
\label{sec: 4-3}

We argue that in addition to temporal patterns of adversarial perturbations, the temporal locations of the perturbations are also important for successful attacks. As a way to check this, we uniformize the attacked videos and obtain the recognition accuracy. In this case, the perturbation in each frame of the uniformized video matches the spatial content in the frame, but is placed in the $T-1$ locations that are different from its original temporal location. Table \ref{table:attack_uniform} compares the accuracy of this case with that before uniformization (i.e., videos that are only attacked). The uniformization process largely neutralizes the effect of the attack, yielding the accuracies quite close to those of the uniformized clean videos (Table \ref{table:clean_uniform}). Therefore, a perturbation at a frame becomes ineffective when it is placed at different temporal locations.

\newcommand{\factorial}{\ensuremath{\mbox{\sc Factorial}}}
\setlength{\textfloatsep}{5pt}
\SetKwInOut{Input}{input}
\SetKwInOut{Output}{output}
\SetKwInOut{Initialization}{initialize}
\SetKwRepeat{Do}{do}{while}
\begin{algorithm}[t]
\footnotesize
\caption{Temporal shuffling for defense}
\label{algorithm}
\Input{attacked video $X_a$, hyperparameters $h_1$ and $h_2$, length of video $T$}
\Output{shuffled video $X_s$}
\Initialization{$t \gets 1$}
\While{$t \leq T$}{
  $l = \max(t-h_1,1)$\\
  $u = \min(t+h_1,T-h_2+1)$\\
  \Do{$t' = t$}{ $t' \gets Rnd(l, u)$ \Comment {\footnotesize{Random integer between $l$ and $u$}}}
  \For{$j \gets 0$ to $h_2-1$ }
  {
    $X_s(t+j) \gets X_a(t'+j)$
  }
  $t \gets t+h_2$}
\Return{$X_s$}
\end{algorithm}
\setlength{\textfloatsep}{\baselineskip}

In addition, when we generate a 3D perturbation by I-FGSM for a uniformized clean video, the perturbation at each frame location is significantly different from each other in spite of the same spatial content, as shown in Figure \ref{fig:sample_pert}. This explains why the attacked and uniformized videos show relatively high accuracy in Table \ref{table:attack_uniform}, and further why adversarial attacks are sensitive to temporal changes.

\section{Temporal shuffling for defense}
\label{sec: 5}

In the previous sections, we found that action recognition models are robust against temporal changes of videos, but adversarial perturbations do not work when the temporal orders are changed.
Based on these findings, we suggest a novel defense method using temporal shuffling.

\begin{figure}[t]
	\centering
	\includegraphics[width=\columnwidth]{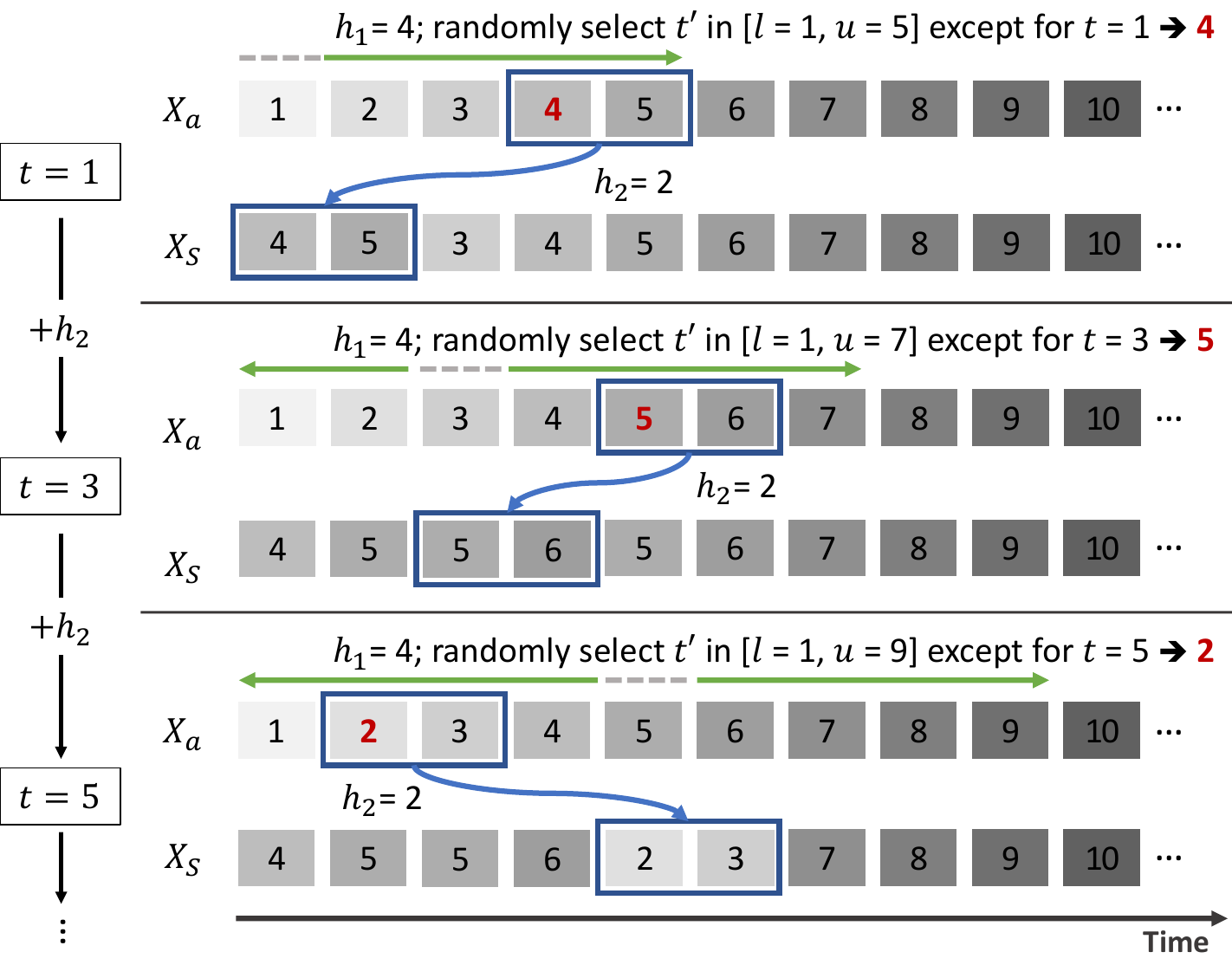}
	\caption{Example of applying Algorithm \ref{algorithm} for $h_1$=4 and $h_2$=2. The first part of the procedure is shown, covering the steps from $t$=1 to $t$=5.}
	\label{fig:algorithm}
\end{figure}

\begin{figure*}[t]
    \centering
	\includegraphics[clip, trim=0cm 8.7cm 0cm 0cm, width=0.95\textwidth]{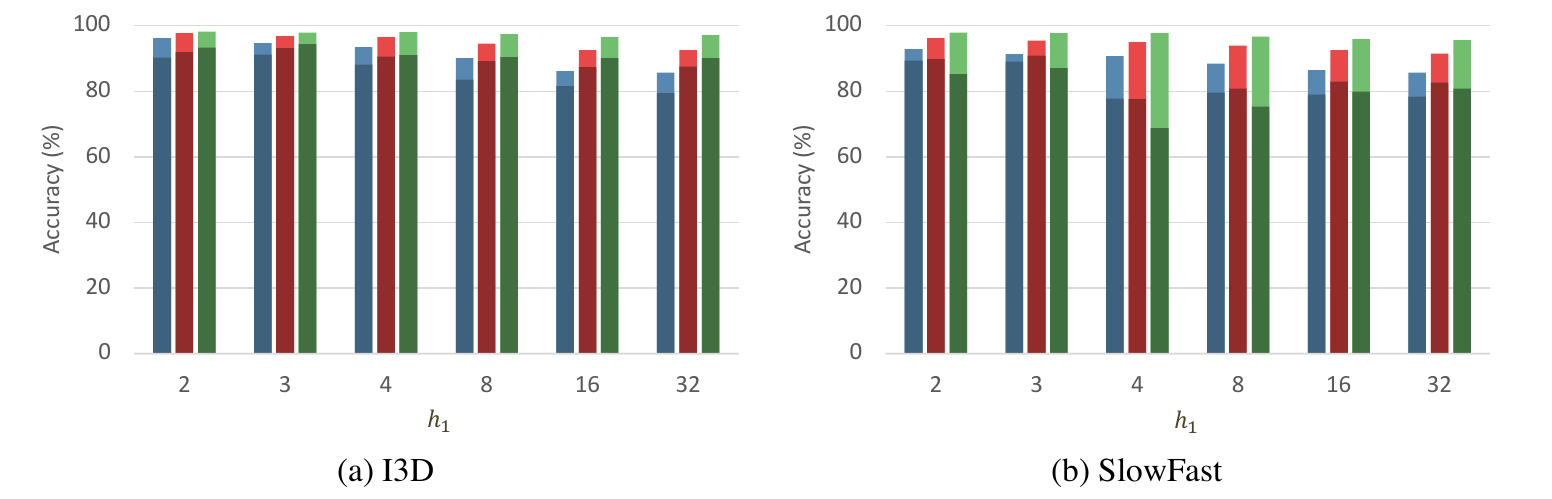}
        \includegraphics[clip, trim=0cm 0cm 0cm 8.7cm, width=0.95\textwidth]{./fig/all_all.pdf}
	\caption{Classification accuracy of our defense method with various combinations of hyperparameters ($h_1$ and $h_2$). The x-axis means the value of $h_1$, and each color (blue, red, or green) indicates a value of $h_2$.}
	\label{fig:all_defense}
\end{figure*}

\begin{figure}[h]
	\centering
	\includegraphics[width=0.8\columnwidth]{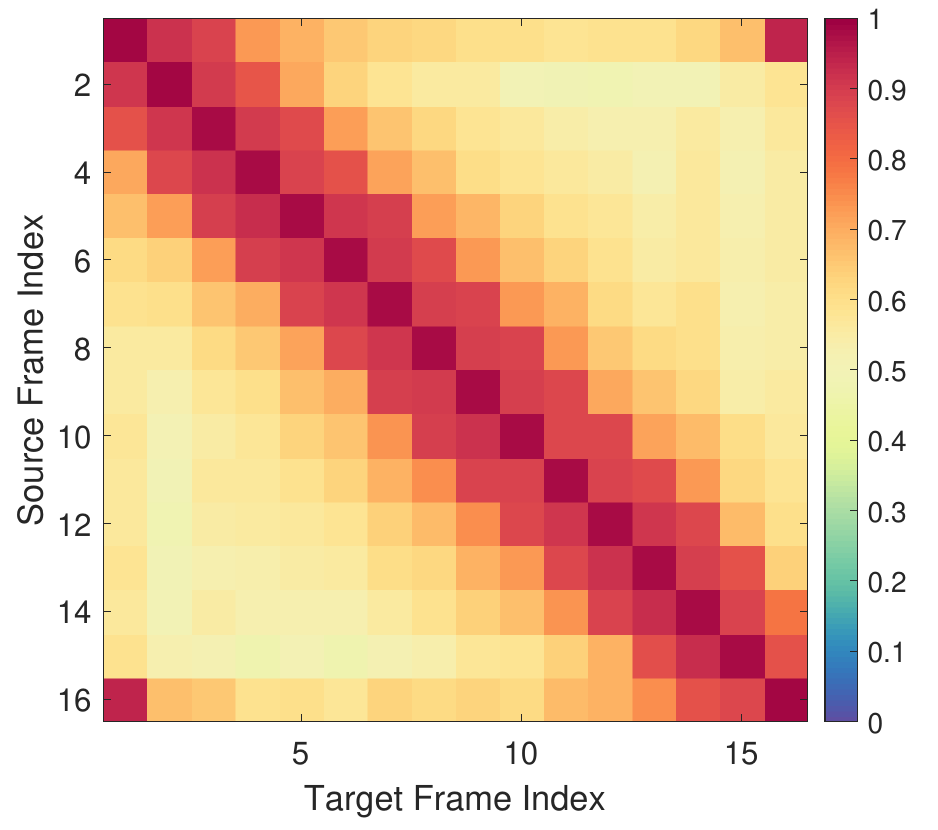}
	\caption{Transferability between different frames in X3D. Each cell (x, y) shows the fooling rate when the x-th frame is attacked using the perturbation generated on the y-th frame. Red colors represent high transferability.}
	\label{fig:x3d_trans}
 \vspace{-1em}
\end{figure}

\subsection{Method}
\label{sec: 5-1}

\newcolumntype{b}{>{\hsize=1\hsize\centering\arraybackslash}X}
\newcolumntype{s}{>{\hsize=1\hsize \centering\arraybackslash}X}
\renewcommand{\arraystretch}{0.85}
\begin{table*}[m]
	\centering
        \caption{Classification accuracy of randomized smoothing (`RS') and combined method (`Ours+RS') with various values of $\sigma_{RS}$. The chosen results are highlighted with bold.}
        \begin{tabularx}{0.9\textwidth}{bbb|sssss}
		\toprule
		\multicolumn{3}{c|}{}&0.008&0.016&0.03&0.06&0.12\\
		\midrule
		\multirow{4}{*}{I3D}&\multirow{2}{*}{RS}&Clean&99.8\%&98.5\%&92.9\%&\textbf{77.9\%}&43.1\%\\
		&&Attacked&0.3\%&0.7\%&15.0\%&\textbf{44.4\%}&36.9\%\\
		\cmidrule{2-8}
		&\multirow{2}{*}{Ours+RS}&Clean&\textbf{97.9\%}&96.8\%&92.2\%&76.7\%&41.5\%\\
		&&Attacked&\textbf{94.4\%}&94.3\%&90.7\%&75.2\%&41.1\%\\
		\midrule
		\multirow{4}{*}{{\shortstack[lb]{SlowFast}}}&\multirow{2}{*}{RS}&Clean&99.7\%&98.6\%&93.7\%&\textbf{77.8\%}&49.8\%\\
		&&Attacked&2.3\%&2.7\%&15.8\%&\textbf{45.7\%}&40.7\%\\
		\cmidrule{2-8}
		&\multirow{2}{*}{Ours+RS}&Clean&\textbf{96.4\%}&95.5\%&90.1\%&75.1\%&44.8\%\\
		&&Attacked&\textbf{90.1\%}&90.9\%&85.7\%&71.6\%&43.6\%\\
		\midrule
		\multirow{4}{*}{\shortstack[lb]{ir-CSN}}&\multirow{2}{*}{RS}&Clean&99.6\%&98.1\%&92.3\%&\textbf{75.8\%}&39.4\%\\
		&&Attacked&8.3\%&9.0\%&26.9\%&\textbf{51.7\%}&35.8\%\\
		\cmidrule{2-8}
		&\multirow{2}{*}{Ours+RS}&Clean&88.1\%&\textbf{87.2\%}&83.1\%&69.5\%&36.7\%\\
		&&Attacked&73.9\%&\textbf{76.4\%}&76.0\%&65.5\%&35.8\%\\
		\midrule
		\multirow{4}{*}{X3D}&\multirow{2}{*}{RS}&Clean&99.2\%&97.5\%&{92.5\%}&\textbf{80.9\%}&47.1\%\\
		&&Attacked&1.8\%&2.6\%&{29.4\%}&\textbf{63.0\%}&44.1\%\\
		\cmidrule{2-8}
		&\multirow{2}{*}{Ours+RS}&Clean&90.6\%&88.7\%&\textbf{83.9\%}&{70.2\%}&38.1\%\\
		&&Attacked&46.2\%&59.4\%&\textbf{70.9\%}&{65.2\%}&36.8\%\\
		\bottomrule
	\end{tabularx}
 	
	\label{table:RSmore}
	\centering
	\vspace{-1.2em}
\end{table*}

\newcolumntype{b}{>{\hsize=1\hsize\centering\arraybackslash}X}
\newcolumntype{s}{>{\hsize=1\hsize \centering\arraybackslash}X}
\renewcommand{\arraystretch}{0.85}
\begin{table*}[m]
	
  	\caption{Classification accuaracy of denoised smoothing (`Denoise') and combined method (`Ours+D') with various values of $\sigma_D$. The chosen results are highlighted with bold.}
   \centering
	\begin{tabularx}{0.7\textwidth}{bbb|sss}
		\toprule
		\multicolumn{3}{c|}{}&0.06&0.12&0.25\\
		\midrule
		\multirow{4}{*}{I3D}&\multirow{2}{*}{Denoise}&Clean&99.4\%&98.5\%&\textbf{88.3\%}\\
		&&Attacked&0.4\%&3.0\%&\textbf{60.1\%}\\
		\cmidrule{2-6}
		&\multirow{2}{*}{Ours+D}&Clean&97.8\%&\textbf{97.2\%}&87.6\%\\
		&&Attacked&95.4\%&\textbf{96.2\%}&87.0\%\\
		\midrule
		\multirow{4}{*}{{\shortstack[lb]{SlowFast}}}&\multirow{2}{*}{Denoise}&Clean&99.6\%&98.9\%&\textbf{92.2\%}\\
		&&Attacked&2.3\%&2.9\%&\textbf{51.5\%}\\
		\cmidrule{2-6}
		&\multirow{2}{*}{Ours+D}&Clean&\textbf{95.8\%}&95.4\%&88.3\%\\
		&&Attacked&\textbf{90.2\%}&90.1\%&84.5\%\\
		\midrule
		\multirow{4}{*}{\shortstack[lb]{ir-CSN}}&\multirow{2}{*}{Denoise}&Clean&99.4\%&98.4\%&\textbf{89.3\%}\\
		&&Attacked&8.3\%&10.7\%&\textbf{67.6\%}\\
		\cmidrule{2-6}
		&\multirow{2}{*}{Ours+D}&Clean&88.5\%&\textbf{87.1\%}&76.6\%\\
		&&Attacked&74.9\%&\textbf{78.8\%}&73.7\%\\
		\midrule    
		\multirow{4}{*}{X3D}&\multirow{2}{*}{Denoise}&Clean&99.6\%&98.7\%&\textbf{90.1\%}\\
		&&Attacked&1.8\%&4.7\%&\textbf{74.4\%}\\
		\cmidrule{2-6}
		&\multirow{2}{*}{Ours+D}&Clean&90.5\%&{89.2\%}&\textbf{79.3\%}\\
		&&Attacked&49.9\%&{67.9\%}&\textbf{76.1\%}\\
		\bottomrule
	\end{tabularx}

	\label{table:Dmore}
	\centering
 	\vspace{-1.5em}
\end{table*}

\newcolumntype{b}{>{\hsize=1.2\hsize\centering\arraybackslash}X}
\newcolumntype{s}{>{\hsize=0.9\hsize \centering\arraybackslash}X}
\renewcommand{\arraystretch}{0.9}
\begin{table}[h]
	\centering
 	\caption{Classification accuracy of our defense method using different ensemble sizes.}

	\begin{tabularx}{0.95\columnwidth}{bb|ssss}
		\toprule
		\multicolumn{2}{c|}{Ensemble size}&1&10&20&100 \\
		\midrule
		\multirow{2}{*}{I3D}&Clean&96.5\%&97.7\%&97.9\%&97.9\%\\
		&Attacked&92.2\%&94.1\%&94.3\%&94.4\% \\
		\midrule
		\multirow{2}{*}{SlowFast}&Clean&95.3\%&95.9\%&96.1\%&96.2\%\\
		&Attacked&87.8\%&89.1\%&89.6\%&89.8\% \\
		\midrule
		\multirow{2}{*}{ir-CSN}&Clean&85.9\%&88.1\%&88.8\%&88.9\%\\
		&Attacked&67.8\%&70.3\%&71.6\%&71.8\% \\
		\midrule
		\multirow{2}{*}{X3D}&Clean&88.6\%&90.8\%&91.1\%&91.3\%\\
		&Attacked&37.5\%&37.6\%&38.0\%&37.3\% \\
		\bottomrule
	\end{tabularx}
	\label{table:ensemble}
	\centering
\end{table}

We design our defense method based on the chunk-based randomization method with a few modifications. First, symmetric randomization of the frame order is ensured, because the chunk-based randomization is asymmetric in that the frame at the chunk boundary can be moved only in the direction keeping it within the chunk. Second, we allow a group of consecutive frames to keep their order in order to maintain the local motion pattern in the group.

Algorithm \ref{algorithm} and Figure \ref{fig:algorithm} show the procedure of our defense method.
First, at each frame location $t$ of the shuffled (i.e., defended) video, we randomly choose a new frame location among $[t-h_1,..., t-1, t+1,...,t+h_1]$.
Then, consecutive $h_2$ frames starting at the chosen frame location in the given (attacked) video are copied.
For instance, when $t$=10, $h_1$=4, and $h_2$=3, the 10th frame of the shuffled video is chosen between the 6th and 14th frames of the given video; if the 14th frame is chosen randomly, the 14th to 16th frames of the given video become the 10th to 12th frames of the shuffled video.

For stability and performance, we use the ensemble technique, i.e., temporal shuffling is conducted multiple times and the final recognition result is obtained by ensembling them \citep{liu2018towards,cao2017mitigating,cohen2019certified,salman2020denoised}.

\subsection{Experimental results}
\label{sec: 5-2}

We evaluate our defense method against various attack methods and compare it to other defense methods. Before evaluation, we find proper hyperparameters and ensemble size not only for our method but also for other defense methods. Then, we measure the performance of our method in various settings of adversarial attack using the found hyperparameters.

\subsubsection{Experimental environments}
The experiments are run on NVIDIA V100 single GPU. All environments are built on the Nvidia-docker and the details of the main libraries are as follows: CUDA 11.0, cuDNN 8, python 3.8, PyTorch 1.7.1, and Torchvision 0.8.2.

\newcolumntype{b}{>{\hsize=1\hsize\centering\arraybackslash}X}
\newcolumntype{h}{>{\hsize=1\hsize\centering\arraybackslash}X}
\newcolumntype{s}{>{\hsize=1\hsize \centering\arraybackslash}X}
\renewcommand{\arraystretch}{0.9}

\begin{table*}[h]
	\centering
        \caption{Evaluation of our method in terms of classification accuracy (\%) against various attack methods (I-FGSM, flickering attack (Flick), and one frame attack (OFA)) in comparison to existing defense methods (randomized smoothing (RS) and denoised smoothing (D)). `No' means the case without defense. `I-FGSM+EOT,' `Flick+EOT,' and `OFA+EOT' represent the three attacks combined with EOT, respectively.
    `+RS' and `+D' indicate the cases where our defense is combined with the randomized smoothing and denoised smoothing, respectively. 
    The best defense method among randomized smoothing, denoised smoothing, and ours is highlighted with bold faces for each attack method.
    The better method between `+RS' and `+D' is underlined when the accuracy is higher than the best defense method.}
	\begin{tabularx}{0.9\textwidth}{bs|s|hss|hss}
		\toprule
        &&Clean&I-FGSM
        &Flickering
        &OFA
        &I-FGSM +EOT
        &Flickering +EOT&OFA +EOT
        \\
		\midrule
		\multirow{6}{*}{I3D} 
            &{No}&100&0.3&29.3&0.4&0.3&29.3&0.4\\
		&RS&77.9&44.4&34.8&42.5&0.3&29.4&8.3\\
		&D&88.3&60.1&36.8&63.3&0.2&34.8&0.8\\                   &Ours&\textbf{97.7}&\textbf{94.1}&\textbf{82.8}&\textbf{97.7}&\textbf{0.6}&\textbf{59.2}&\textbf{96.9}\\
            \cmidrule{2-9}
		&+RS&\underline{97.9}&94.4&82.4&97.4&0.6&58.8&\underline{97.4}\\
		&+D&97.2&\underline{96.2}&81.5&97.1&\underline{}{0.8}&56.4&97.0\\
		\midrule
		\multirow{6}{*}{\shortstack[lb]{SlowFast}} 
            &{No}&100&2.3&36.1&3.3&2.3&36.1&3.3\\
		&RS&77.8&45.7&40.9&55.5&1.1&27.1&20.0\\
		&D&92.2&51.5&46.4&74.8&1.3&41.3&4.2\\
		  &Ours&\textbf{95.9}&\textbf{89.1}&\textbf{80.4}&\textbf{95.4}&\textbf{2.6}&\textbf{61.2}&\textbf{85.8}\\
            \cmidrule{2-9}
            &+RS&\underline{96.4}&90.1&80.4&95.3&0.6&59.6&86.3\\
		&+D&95.8&\underline{90.2}&\underline{80.7}&\underline{95.7}&3.8&56.2&\underline{87.0}\\
		\midrule
		\multirow{6}{*}{\shortstack[lb]{ir-CSN}} 
            &{No}&100&8.3&31.9&9.9&8.3&31.9&9.9\\
		&RS&75.8&51.7&39.3&51.7&5.1&26.6&25.0\\
		&D&\textbf{89.3}&67.6&40.8&74.3&5.7&33.1&9.9\\
		&Ours&88.1&\textbf{70.3}&\textbf{71.2}&\textbf{86.5}&\textbf{6.2}&\textbf{54.7}&\textbf{82.0}\\
        \cmidrule{2-9}	&+RS&87.2&76.4&\underline{71.6}&85.3&\underline{6.4}&50.9&\underline{83.4}\\
        &+D&87.1&\underline{78.8}&70.2&86.1&5.9&48.0&83.1\\
		\midrule
		\multirow{6}{*}{X3D} 
        &{No}&100&1.8&36.7&1.7&1.8&36.7&1.7\\
		&RS&80.9&63.0&51.6&62.3&1.2&33.7&\textbf{27.4}\\
		&D&90.1&\textbf{74.4}&56.3&\textbf{81.3}&\textbf{1.3}&40.4&2.5\\
		&Ours&\textbf{90.8}&37.6&\textbf{73.3}&48.3&1.1&\textbf{58.5}&16.9\\ \cmidrule{2-9}
		&+RS&83.9&70.9&68.3&73.4&\underline{1.3}&46.5&\underline{42.5}\\
        &+D&79.3&\underline{76.1}&64.5&77.7&0.7&38.8&14.9\\
		\bottomrule
	\end{tabularx}
	\label{table:all}
\end{table*}

\subsubsection{Hyperparameters and ensemble size}
\label{sec: 5-2-2}
\vspace{0.5em}

\noindent \textbf{Hyperparameters for our method.} Figure \ref{fig:all_defense} shows the accuracy of our defense method against I-FGSM with $\epsilon$=4/255 for various combinations of the hyperparameters. The ensemble size is set to 100 to consider a sufficiently large ensemble. 
Considering the performance for both clean videos and attacked videos, we can choose an optimal hyperparameter set as $\{h_1, h_2\}=\{3,4\}$ for I3D, $\{2, 2\}$ for SlowFast, $\{16,2\}$ for ir-CSN, and $\{16,1\}$ for X3D.
\citeauthor{hwang2021just} showed that ir-CSN has high transferability of perturbations between nearby frames, while I3D and SlowFast has low transferability. We additionally found that X3D has high transferability as shown in Figure \ref{fig:x3d_trans}. 
Thus, ir-CSN and X3D need large values of $h_1$ to avoid shuffling among nearby frames.
In the case of SlowFast, it has high transferability between every four frame locations \citep{hwang2021just}.
Thus, setting $h_1<4$ is effective for defense to exclude possibility of copying frames between transferable frame locations.
If $h_2$ is large, local motion patterns tend to be preserved.
As a result, the accuracy of clean videos is kept high, but the accuracy of attacked videos remains low in many cases because temporal patterns in the perturbations are not destroyed well.
A reasonable choice of the value of $h_2$ to balance this trade-off varies among the models.
\\[-0.5\baselineskip]

\newcolumntype{b}{>{\hsize=1\hsize\centering\arraybackslash}X}
\newcolumntype{s}{>{\hsize=1\hsize \centering\arraybackslash}X}
\renewcommand{\arraystretch}{0.9}
\begin{table}[h]
	\centering
         \caption{Classification accuracy of our defense method on I-FGSM and one frame attack (`OFA') with various values of $\epsilon$.}
         \label{table:more}
	\begin{tabularx}{\columnwidth}{bb|sss}
		\toprule
		&&\multicolumn{2}{c}{I-FGSM}&OFA \\
		&&$\epsilon$=8/255&$\epsilon$=16/255&$\epsilon$=16/255\\
		\midrule
		\multirow{5}{*}{I3D}&No&0.2\%&0.1\%&0.4\%\\
		&RS&29.4\%&12.1\%&43.0\% \\
		&Denoise&46.4\%&32.2\%&62.8\% \\
		&Ours&92.0\%&89.0\%&97.6\% \\
		&Ours+RS&92.8\%&89.3\%&97.5\% \\
		&Ours+D&94.9\%&92.8\%&97.0\% \\
		\midrule
		\multirow{5}{*}{SlowFast}&No&2.3\%&2.2\%&3.2\%\\
		&RS&29.7\%&13.2\%&55.5\% \\
		&Denoise&36.3\%&24.5\%&74.8\% \\
		&Ours&87.2\%&84.3\%&95.4\% \\
		&Ours+RS&87.5\%&83.4\%&95.4\% \\
		&Ours+D&87.3\%&83.6\%&95.1\% \\
		\midrule
		\multirow{5}{*}{ir-CSN}&No&8.1\%&7.7\%&9.9\%\\
		&RS&36.5\%&21.5\%&51.3\% \\
		&Denoise&52.8\%&40.7\%&74.1\% \\
		&Ours&62.2\%&49.6\%&87.0\% \\
		&Ours+RS&67.8\%&55.3\%&85.5\% \\
		&Ours+D&73.2\%&63.3\%&86.0\% \\
		\midrule
		\multirow{5}{*}{X3D}&No&1.7\%&1.6\%&1.6\%\\
		&RS&50.6\%&29.5\%&61.9\% \\
		&Denoise&64.4\%&54.1\%&81.3\% \\
		&Ours&21.1\%&10.3\%&47.8\% \\
		&Ours+RS&60.6\%&41.9\%&74.2\% \\
		&Ours+D&72.1\%&68.4\%&77.6\% \\
		\bottomrule
	\end{tabularx}
	
	\centering

\end{table}

\newcolumntype{b}{>{\hsize=0.6\hsize\centering\arraybackslash}X}
\newcolumntype{s}{>{\hsize=1.2\hsize \centering\arraybackslash}X}
\renewcommand{\arraystretch}{1}
\begin{table}[h]
	\centering
         \caption{Classification accuracy of the combined defense method using our method with various values of the hyperparameters and an existing defense method (randomized smoothing (`Ours+RS') or denoised smoothing (`Ours+D')) in the case of X3D. $\sigma_{RS}$=0.03 and $\sigma_D$=0.25 are used. The performance reported in Table \ref{table:all} is underlined. The best defense performance in each column is highlighted in bold.}
	
	\begin{tabularx}{\columnwidth}{bb|ss|ss}
		\toprule
		\multirow{2}{*}{$h_1$}&\multirow{2}{*}{$h_2$}&\multicolumn{2}{c|}{Ours+RS}&\multicolumn{2}{c}{Ours+D}\\
		&&Clean&Attacked&Clean&Attacked\\
		\midrule
		\multirow{3}{*}{2}&1&86.7\%&67.1\%&84.5\%&78.4\%\\
		&2&89.7\%&60.6\%&88.6\%&79.6\%\\
		&4&\textbf{91.1\%}&52.6\%&87.9\%&79.3\%\\
		\midrule
		\multirow{3}{*}{3}&1&85.7\%&69.5\%&83.8\%&78.7\%\\
		&2&89.3\%&63.6\%&86.5\%&\textbf{80.4\%}\\
		&4&{90.6\%}&53.9\%&87.9\%&79.5\%\\
		\midrule
		\multirow{3}{*}{4}&1&85.4\%&70.6\%&82.2\%&78.3\%\\
		&2&85.4\%&70.6\%&85.7\%&79.1\%\\
		&4&{90.6\%}&55.7\%&\textbf{88.0\%}&79.1\%\\
		\midrule
		\multirow{3}{*}{8}&1&84.4\%&70.8\%&79.9\%&75.7\%\\
		&2&87.7\%&66.8\%&84.9\%&79.6\%\\
		&4&89.9\%&57.9\%&87.3\%&79.8\%\\
		\midrule
		\multirow{3}{*}{16}&1&\underline{{83.9\%}}&\underline{\textbf{70.9\%}}&\underline{79.3\%}&\underline{76.1\%}\\
		&2&87.6\%&66.7\%&85.4\%&80.3\%\\ 
		&4&89.7\%&58.0\%&87.0\%&79.3\%\\
		\bottomrule
	\end{tabularx}
 \label{table:x3ds}
	\centering

\end{table}

\noindent \textbf{Hyperparameters for compared defense methods.}
We compare our method to two representative defense methods used in object recognition: randomized smoothing for certified robustness \citep{cohen2019certified} and denoised smoothing for provable defense \citep{salman2020denoised}.
Furthermore, we examine the performance when our method is combined with one of them; our method destroys the temporal structure of perturbations while randomized smoothing and denoised smoothing compensate for spatial perturbations, and thus performance boosting may be expected by combining them. Note that we do not apply additional training to pre-trained action recognition models in order to compare different defense methods fairly.

To determine their hyperparameters, we follow a procedure similar to that determining the hyperparameters of our method.
In other words, we conduct experiments using various values of the hyperparameters under I-FGSM ($\epsilon$=4/255) and the best performing values are chosen, which is detailed below. 

In the case of randomized smoothing, the hyperparameter $\sigma_{RS}$ determines the standard deviation of the Gaussian random noise added to the input data for defense. 
We try $\sigma_{RS} \in \{0.008, 0.016, 0.03, 0.06, 0.12\}$, which are smaller than those used in the original paper by considering the higher dimension of the input data in our case. 
The results for different values of $\sigma_{RS}$ are shown in Table \ref{table:RSmore}. 
When the randomized smoothing is used alone, the value of $\sigma_{RS}$ yields a trade-off relationship between the performance on the clean videos and that on the attacked videos.
Considering this, we choose $\sigma_{RS}$=0.06.
When the randomized smoothing is used in combination with our method, we choose the best value for each model, i.e., 0.008 for I3D and SlowFast, 0.016 for ir-CSN, and 0.03 for X3D.

In the case of denoised smoothing, the hyperparameter $\sigma_{D}$ determines the standard deviation of the Gaussian random noise added to the training data of the denoiser. 
We set $\sigma_D \in \{0.06, 0.12, 0.25\}$, and the corresponding results are shown in Table \ref{table:Dmore}.
We choose $\sigma_D$=0.25 when the denoised smoothing is used alone, and $\sigma_D$=0.06 for SlowFast, 0.12 for I3D and ir-CSN, and 0.25 for X3D.
\\[-0.5\baselineskip]

\noindent \textbf{Ensemble size.} With these optimal hyperparameter sets, Table \ref{table:ensemble} shows the classification accuracy with respect to the ensemble size.
While a larger size of ensemble yields higher accuracy, the improvement is marginal. Thus, we set the ensemble size to 10 in the following comparison study.

\subsubsection{Comparison} 
Our defense method is evaluated against several attack methods in comparison to the two existing defense methods (randomized smoothing and denoised smoothing) as mentioned above. 

For attack methods, we use I-FGSM \citep{kurakin2016adversarial} (the 3D version explained in the previous section) with $\epsilon$=4/255, flickering attack \citep{pony2021over}, and one frame attack \citep{hwang2021just} with $\epsilon$=8/255. 
The flickering attack and one frame attack specifically target action recognition models.
For evaluating against adaptive attack, we also test the cases where the expectation over transformation (EOT) technique \citep{athalye2018obfuscated,athalye2018synthesizing} is combined with each attack method,
which works with knowledge of the existence and type of defense.

Table \ref{table:all} shows the comparison results. Except for a few cases of X3D, our method achieves higher accuracy than the existing defense methods.
Furthermore, we find that combining our method with the existing defense methods shows further improvement in many cases.
Our method is also effective against the one frame attack. Although a perturbation created by the one frame attack does not have a temporal structure, our defense can neutralize it by changing its temporal location.
Our method is more effective than the other defense methods when EOT is applied, especially for the attacks targeting action recognition models.
On the other hand, defense against I-FGSM combined with EOT remains challenging (similar observations were made by \citeauthor{athalye2018obfuscated} and \citeauthor{adaptive}), which would need further research in the future.

Table \ref{table:more} provides the results with larger values of $\epsilon$ compared to Table \ref{table:all}.
The effectiveness of our method (used alone or combined with an existing defense method) is also verified in these results.

When our method and one of the existing defense methods are combined, we optimize the hyperparameters of the latter, and the hyperparameters of our method are fixed as explained in Section \ref{sec: 5-2-2}. However, we find the possibility of improved performance by additional optimization of the hyperparameters of our method, particularly for X3D. Table \ref{table:x3ds} shows the performance for various values of the hyperparameters of our method when the videos are attacked by I-FGSM ($\epsilon$=4/255). We can observe that our method combined with an existing method can become more powerful.

\section{Conclusion}
We suggested that the action recognition models depend largely on spatial information for recognition, leading to an interesting phenomenon that the models are fairly robust to temporal randomization of input videos.
On the other hand, we found that changing temporal orders of adversarial perturbations is fatal to a successful attack.
Based on these discoveries, we proposed a defense method using temporal shuffling for action recognition models.
Our method achieved high defense performance under various attack and defense scenarios without additional training.
In the future, we plan to extend our analysis to other video recognition models and datasets.
\\[-0.1\baselineskip]

\noindent{\large{\textbf{Acknowledgement}}}
\\[-0.7\baselineskip]

This work was supported by the Artificial Intelligence Graduate School Program (Yonsei University, 2020-0-01361).

\printcredits

\bibliographystyle{cas-model2-names}

\bibliography{main}



\end{document}